\pdfoutput=1

\documentclass[11pt]{article}

\usepackage[]{acl}

\usepackage{times}
\usepackage{latexsym}
\usepackage{float}

\usepackage{amsmath}

\usepackage{enumitem}  
\usepackage[capitalise,noabbrev]{cleveref} 

\usepackage[T1]{fontenc}

\usepackage[utf8]{inputenc}

\usepackage{graphicx}

\usepackage[boxed]{algorithm2e}

\usepackage{microtype}

\usepackage{inconsolata}

\usepackage{authblk}

\usepackage{booktabs}
\newcommand{\vsa}[1][4pt]{\mathrel{%
   \hbox{\rule[\dimexpr\fontdimen22\textfont2-.2pt\relax]{#1}{.4pt}}%
   \mkern-4mu\hbox{\usefont{U}{lasy}{m}{n}\symbol{41}}}}
   
\title{Adapting Large Language Models for Character-based Augmentative and Alternative Communication}

\author[1,2*]{Dylan Gaines}
\author[1*]{Keith Vertanen}

\affil[1]{Department of Computer Science, Michigan Technological University, Houghton, MI, USA}
\affil[2]{Department of Computer Science, Kennesaw State University, Marietta, GA, USA}
\affil[ ]{\small{\href{mailto:dgaine20@kennesaw.edu}{dgaine20@kennesaw.edu}, \href{mailto:vertanen@mtu.edu}{vertanen@mtu.edu}}}

\begin{document}
\maketitle

\def\thefootnote{*}\footnotetext{These authors contributed equally to this work.}
\def\thefootnote{\arabic{footnote}}

\begin{abstract}
Users of Augmentative and Alternative Communication (AAC) may write letter-by-letter via an interface that uses a character language model. However, most state-of-the-art large pretrained language models predict subword tokens of variable length. We investigate how to practically use such models to make accurate and efficient character predictions. Our algorithm for producing character predictions from a subword large language model (LLM) provides more accurate predictions than using a classification layer, a byte-level LLM, or an n-gram model. Additionally, we investigate a domain adaptation procedure based on a large dataset of sentences we curated based on scoring how useful each sentence might be for spoken or written AAC communication. We find our procedure further improves model performance on simple, conversational text. 


\end{abstract}

\section{Introduction}
Augmentative and Alternative Communication (AAC) devices allow non-speaking individuals to communicate in face-to-face conversations and asynchronously via messaging systems such as email. Depending on their disability, AAC users may be unable to use a conventional computer keyboard and may instead rely on alternative input methods such as eye tracking or triggering a single switch (e.g.~by twitching a muscle or puffing air). Unfortunately, such input methods can make writing slow and tiring. The writing speed of non-speaking AAC users varies, but is frequently less than 10 words per minute (WPM), with an average closer to 5 WPM~\cite{polacek-text}. 

While some AAC users may engage in long conversations, communications are often shorter and more isolated in nature due to the input rate limitation. AAC users with severe motor impairments may write letter-by-letter using an interface such as Dasher \cite{ward_dasher}, the RSVP Keyboard~\cite{orhan_rsvp}, or Nomon \cite{broderick_nomon}. These AAC interfaces all require the probability distribution over the next character based on a user's previous text. 

We wanted to know if recent advances in pretrained large language models (LLMs) really help with the ``simple'' case of predicting the next letter in a sentence with only that sentence as context. Numerous papers have shown that ever-larger models such as Llama 3~\cite{grattafiori_llama3}, Mistral 7B~\cite{jiang_mistral_7b_2023}, and GPT-4~\cite{OpenAI_gpt4_2024} provide gains on more ``complex'' natural language processing tasks such as question answering, sentiment analysis, and machine translation. 

Many large transformer models do not operate on a character level. Rather, they use a fixed vocabulary of subword tokens when making their predictions. These subword tokens can range anywhere from a single character to a full word in length. These tokens are learned during the training process and stem from frequently seen sequences of characters. As we detail later, this subword tokenization makes it difficult to obtain single character predictions from many LLMs. 

In this work we make four contributions: (1) an algorithm that produces a probability distribution over single characters from a subword token language model, (2) offline, novel test sets of AAC-like text that can be used to evaluate LLMs without fear of their inclusion in the models' training data, (3) a classification of existing large (87\,B words) training corpora based on their similarity to AAC-like text, and (4) an empirical evaluation and comparison of different methodologies for obtaining single-character predictions using LLMs. 

We have released our algorithm's source code, our classified data sets, our classification model, and our best language models as part of this work.\footnote{\url{https://osf.io/ajm7t/}}

This paper is structured as follows: first, we motivate why character predictions are important for AAC text entry (\cref{sec_related_work}) and describe our algorithm for making such predictions using a subword tokenized LLM (\cref{sec_search_alg}). Next, we find the best starting model, testing performance on a private collection of AAC-like text (\cref{sec_pretrained}). We then explore adapting the LLM using small in-domain sets and using text classified from much larger out-of-domain sets (\cref{sec_optimizing}). Next, we compare our search algorithm against other possible approaches (\cref{section_other_models}). Finally, we show our results on five test sets (\cref{sec_final_eval}) before wrapping up with discussion (\cref{sec_discussion}) and conclusions (\cref{sec_conclusions}).

\section{Related work\label{sec_related_work}}

\subsection{Language modeling for AAC}
Language models have often been studied in AAC in hopes of improving communication speed. Work has focused on four main topics: word completions and predictions, sentence and phrase retrieval, abbreviation expansion, and domain adaptation.

\textbf{Word completions and predictions.}
Word completions attempt to predict a user's intended current word before they finish typing it, while word predictions attempt to predict subsequent words based on the likelihood of each word in the context of the user's sentence. Prior work by \citet{kristensson_design_and_analysis} showed that word predictions can increase or decrease a user's entry rate depending on how and when predictions are displayed and the user's typing strategy. \citet{yusufali_bridging_2023, yusufali_refining_2024} used LLMs to generate word predictions for AAC devices and showed entry rates up to 30.4 WPM.

\textbf{Sentence and phrase retrieval.}
Retrieval techniques allow users to store sentences or phrases that they frequently use~\cite{todman_whole_2008}. These messages can then be retrieved by the user by entering specific keywords~\cite{langer_using_1998} or matching context tags~\cite{kristensson_design_engineering_2020}. The drawback to retrieval is utterances must be pre-stored by the user in their device, so they are not as adaptable to different conversational situations. 

\textbf{Abbreviation expansion.}
Recent advances in LLMs have allowed the development of more dynamic abbreviation expansion interfaces. \citet{cai_context_aware_2022, cai_speak_faster_2023} developed a system where users abbreviate each word to its first letter. The system leverages the context of the utterance and an LLM to decode the intended text. KWickChat~\cite{shen_kwickchat_2022} allows the user to enter a few keywords and uses an LLM to generate expanded utterances based on the provided keywords, some contextual details about the user, and the dialogue from their conversational partner. \citet{valencia_less_i_type_2023} also developed an abbreviation expansion system that expands single words into complete sentences using conversational context and background information about the user.

\textbf{Domain adaptation.}
The type of text an AAC user may want to write can differ from typical sources of language model training data (e.g.~text from web pages). \citet{vertanen_imagination} used a collection of crowdsourced fictional AAC messages to filter sentences from large social media datasets using cross-entropy difference \cite{moore_intelligent}. Their optimized n-gram language models improved perplexity and potential keystroke savings on AAC-like test sets. \citet{adhikary_abbreviated} used the same crowdsourced collection to select sentences from text scraped from the web. They found using a BERT-based classifier to select sentences slightly outperformed selection via cross-entropy difference. 

\subsection{Applications for character predictions}
While word and phrase-based language modeling approaches can help accelerate input with many AAC devices, some devices rely primarily on character-at-a-time input. In these systems, it may take several user interactions to input a single character due to a user's noisy input signal. Character predictions from a language model can help accelerate input of individual characters in such systems.

\textbf{Brain-computer interfaces.}
P300 spellers~\cite{farwell_talking} present visual stimuli to users and measure their response using EEG electrodes. For example, the RSVP Keyboard~\cite{orhan_rsvp} displays a rapid series of characters to users. Using RSVP, it can take several series worth of EEG responses to infer a user's desired character. RSVP uses a character n-gram language model in combination with the EEG evidence to reduce the number of series needed. We hypothesize an LLM will provide more accurate probabilities and reduce the number of series needed.

\textbf{Nomon.}
Nomon~\cite{broderick_nomon,bonaker_performance_2022,bonaker_demonstrating_2022} allows users to select words or characters by activating a switch when the rotating clock corresponding to their target reaches noon. It can take several clicks to make a single selection since the system needs to differentiate between similar clocks. Since Nomon provides both word and character selection options, it needs a language model that can evaluate the likelihood of each possibility to help accelerate writing.

\textbf{Dasher.}
Dasher~\cite{ward_dasher} inherently relies on character-level language model predictions. In Dasher, the letters of the alphabet are displayed vertically on the right side of the interface. To enter text, users move their cursor through the sequence of nested letter boxes corresponding to their desired text. Dasher can be controlled by various devices such an eye-tracker, mouse, or touchscreen. As users progress through their text, more likely letter options are displayed as larger targets, making it easier to select them.

Dasher uses the prediction by partial match (PPM) algorithm \cite{cleary_ppm}. PPM is similar to an n-gram model but has a data structure making it easy to continually adapt to a user's writing. We hope to improve on this using LLMs. However, a key challenge is generating predictions quickly enough to support Dasher's visualization of the many possible future character sequences. 

\section{Character prediction algorithm\label{sec_search_alg}}

As we discussed previously, many modern large transformer models use subword tokens instead of single characters or full words. The first step in using a subword token language model is to convert the context into a sequence of tokens via tokenization. Each model comes with its own tokenizer, which uses a greedy algorithm, such as byte pair encoding~\cite{sennrich_neural_2016}, to segment any given text into tokens in its model's vocabulary. 

This subword tokenization complicates the process of obtaining a distribution of character probabilities since adding another character to the end of a section of text could (1) begin a new token containing only the new character, (2) extend the existing most recent token with the new character, or (3) rearrange the optimal tokenization of the most recent word. For the last case, take for example writing the word ``yesterday''. If the user has written ``yeste'' thus far, the most likely tokenization may be into ``y-este'' (where - denotes token boundaries). Adding an ``r'' could result in the most likely tokenization changing to ``yes-ter''. 

To properly consider each of these possibilities, we must remove one or more tokens from the end of the context to allow the model the option of extending it, as opposed to only adding a new token. Removing more tokens allows the model more flexibility to rearrange the optimal tokenization and in our experience leads to better predictions, but at the cost of a higher prediction time. In this work, we remove all tokens in the current in-progress word to allow maximal flexibility in tokenizing the partial word. Going further back would provide no benefit because, in all models we used, the space character only appears at the start of tokens and thus tokens cannot span word boundaries.

We then ask the model to estimate the likelihood of the next token and evaluate any token that matches our context. For efficiency, we only track a certain number of hypotheses at a time, known as the \textit{beam width}. We continue expanding each hypothesis until it has regenerated the removed context tokens plus at least one character. We then store the likelihood for each final prediction in a list based on the character that directly follows the context. We continue this process until there are no more hypotheses to extend, or, for efficiency, the number of completed hypotheses exceeds a \textit{max completed} parameter. Then we sum the likelihoods stored for each character in our symbol set and normalize to sum to one, arriving at a distribution over our symbol set. See \cref{appendix_pseudocode} for our pseudocode search algorithm or our supplementary materials for a Python implementation.

To give a concrete example of the benefit of removing tokens from the end of the context, we further examine the ``yeste'' example. If we do not remove any context, and simply marginalize the probabilities over the first character in each token, the probability of the next letter being R is 0.0047. Using our search algorithm with context removal, this jumps to 0.9999. We suspect this is due to the training process of the model; during training it doesn't frequently encounter fragmented words and is allowed to tokenize the full word in the most optimal way. By not removing any context, we force the model to infer text with suboptimal tokenizations, causing it to struggle.


\section{Pretrained model performance\label{sec_pretrained}}
How do pretrained transformer language models perform on AAC-like text? In this section, we focus on LLMs using subword tokenization. 

\subsection{Private AAC-like test data\label{private_test_data}}
Unfortunately, actual text from AAC users is difficult to source due to practical, ethical, and privacy reasons. There are a number of datasets that have been used to approximate AAC communications such as voice banking messages \cite{costello_message}, Switchboard \cite{godfrey_switchboard}, DailyDialog \cite{li_dailydialog}, COMM2 \cite{vertanen_comm2}, and Turk dialogues \cite{vertanen_aacdialogue}. However, these datasets are available online and may have been used in training the LLMs we want to evaluate.

As we wish to fairly compare performance between different LLMs, as well as between LLMs and an n-gram language model optimized for AAC-like text, we created two private tests sets, one for informal written communications (i.e.~similar to mobile messaging), and one for person-to-person communications. We denoted the two distinct genres as \textit{written} and \textit{spoken} respectively.\footnote{We intend to provide access to researchers on the condition they not leak the text online or to hosted models.}  We sourced our text from compositions made by workers on Amazon Mechanical Turk in response to various prompts. For details of how we constructed our test sets, including example sentences, see \cref{appendix_test_set}.

We split each genre in half, creating a written dev set (8.6\,K words), a written test set (9.0\,K words), a spoken dev set (11.3\,K words), and a spoken test set (10.8\,K words). We use the dev sets for most of our experiments, reserving the test sets for a final evaluation (\cref{sec_final_eval}). Our final evaluation also uses text written by an AAC user, messages written by people with ALS for voice banking \cite{costello_message}, and responses to common conversational situations \cite{vertanen_comm2}.
  
\subsection{Perplexity experiments}
We converted our private test sets to lowercase and removed anything not an English letter, apostrophe, or space. We calculated per-character perplexity (including spaces) on our dev sets without any sentence end token and using the sentence start token appropriate for the model under test. 

We used 16-bit floating point inference for the LLMs. We used our search algorithm (\cref{sec_search_alg}) for LLM predictions. We found a search beam of 8 and a max completed stopping criteria of 32\,K provided a perplexity close to the perplexity without pruning while significantly speeding inference.

We compared against a baseline of a 12-gram language model optimized for AAC-like text, denoted \textit{AAC n-gram}.\footnote{\url{https://osf.io/c6mnz}} This model was trained using Witten-Bell smoothing on 21\,B characters from Common Crawl, Reddit, movie subtitles, and Twitter. Training sentences were selected via cross-entropy difference selection \cite{moore_intelligent} using in-domain models trained on DailyDialog \cite{li_dailydialog}, short emails, and AAC-like messages \cite{vertanen_imagination}.

As shown in \cref{table_compare}, per-character perplexity decreased with increasing model size, though gains plateaued for the OPT model family \cite{zhang_opt} at around 1.3\,B parameters. Thus, even given the ``easy'' nature of our task (i.e.~predicting the next character within an isolated sentence), bigger models were often better. While the n-gram model optimized for AAC did better than DistilGPT with a similar number of parameters, larger LLMs eventually outperformed it despite the LLMs not being fine-tuned on AAC-like text. 

The OPT models outperformed the GPT-2 models \cite{radford_gpt2} at similar model sizes. We also tested the recent Llama 3 model \cite{grattafiori_llama3} that was trained on substantially more data than the other models (15\,T tokens), but found it did no better than the OPT models. 

Determining the cause of the perplexity differences between model families is difficult due to the many differences in the underlying LLM architectures, training data, and training procedures. But given that the performance of n-gram models on AAC-like text have been shown to depend on the training data \cite{vertanen_imagination}, we conjecture the difference may be due to the amount of well-matched English text versus other types of text (e.g.~other natural or programming languages).  

Our n-gram model inferences using KenLM \cite{heafield_kenlm} were fast at 0.1\,ms per prediction (using a CPU). Despite substantial performance optimization efforts as well as using an NVIDIA A100 GPU for inference, our search algorithm was much slower at 55--120\,ms per prediction. In general, larger models took longer. We will investigate other methods for faster inference in \cref{section_other_models}.

\begin{table}[tb]
\setlength{\tabcolsep}{4pt}
\centering
\small
\begin{tabular}{ l r r r }
\toprule
                  & Params (M) & Perplexity & Time (ms) \\
\midrule
AAC n-gram & 70 & 2.55 & 0.1 \\
\midrule
distilgpt2 & 82 & 2.71 & 55.0 \\
gpt2 & 124 & 2.54 & 60.0 \\
gpt2-medium & 355 & 2.45 & 70.1 \\
gpt2-xl & 1558 & 2.38 & 91.7 \\
\midrule
opt-125m & 125 & 2.39 & 64.3 \\
opt-350m & 331 & 2.31 & 78.6 \\
opt-1.3b & 1316 & 2.22 & 77.9 \\
opt-2.7b & 2652 & 2.21 & 86.8 \\
opt-6.7b & 6658 & 2.20 & 89.5 \\
opt-13b & 12853 & 2.23 & 108.8 \\
\midrule
Meta-Llama-3.1-8B & 8030 & 2.31 & 119.9 \\
Meta-Llama-3.2-1B & 1236 & 2.37 & 97.0 \\
Meta-Llama-3.2-3B & 3213 & 2.33 & 115.0 \\
\bottomrule
\end{tabular}
\caption{Model parameters, per-character perplexity, and prediction time of pretrained LLMs on our dev sets (average of the written and spoken sets).\label{table_compare}}
\end{table}

\section{Optimizing language models for AAC\label{sec_optimizing}}
The LLMs in the previous section were all trained on general text from the web. As with previous work on n-gram language models \cite{vertanen_imagination,adhikary_abbreviated}, we may be able to improve predictions on AAC-like text by focusing the model on this style of text. We selected the opt-350m model for further refinement as it provided the majority of perplexity gains while being small enough to plausibly be used on an end-user's device. For privacy and latency reasons, users may prefer to use a local model. 

Our results in \cref{table_compare} were on lowercase versions of our test sentences. Our search algorithm sums over all matching following tokens ignoring case. However, we conditioned  predictions on all lowercase context. This simulates an AAC interface where the user does not bother to denote case, or when casing is added later in some way. We found we could slightly improve perplexity on our dev sets using opt-350m from 2.31 to 2.30 by upper casing the first letter of the sentence and the words: I, I'm, I'll, I've, and I'd. We found this simple approach performed similar to using the human-supplied case. We use this automatic casing method for the remainder of our experiments. 

\subsection{In-domain datasets}
To approximate written AAC, we used sentences from forum posts made by people using a mobile device \cite{vertanen_mining}. We denote this dataset as \textit{Forum}. Our rationale for using this dataset was that it contains a rich set of topics while also having a bias towards concise writing. 

To approximate spoken AAC, we used the DailyDialog corpus \cite{li_dailydialog} and the BOLT SMS/Chat corpus \cite{song_bolt}. We denote these datasets as \textit{Daily} and \textit{BOLT} respectively. Our rationale for choosing these datasets was that while the text was actually written and not spoken, it focuses on person-to-person dialogues. Compared to other datasets based on spoken dialogues such as Switchboard \cite{godfrey_switchboard}, the text is relatively clean and does not contain spoken speech artifacts. Daily had 46\,K sentences (356\,K words), BOLT had 271\,K sentences (1.9\,M words), and Forum had 888\,K sentences (10.1\,M words). 

\subsection{Large datasets}
We selected two large corpora to mine data from: the Colossal Clean Crawled Corpus \cite{raffel_t5} (ODC-By License) and the OpenSubtitles2016 movie subtitle corpus \cite{lison_subtitles} (License unknown). We denote these datasets as \textit{C4} and \textit{Subtitle} respectively. This selection was guided by results\footnote{\url{https://osf.io/ajm7t/wiki/lm_char_dec19/}} showing that out of 14 possible training sources, these two corpora performed the best on modeling AAC-like text. Further, we think C4 provides text that well-approximates a written style of communication while Subtitle represents a more spoken style.
 
We filtered C4 and Subtitle to sentences that started with a capital letter, had both uppercase and lowercase characters, and ended with sentence end punctuation. We dropped sentences where more than 20\% of the words were out-of-vocabulary with respect to a list of 860\,K words obtained from human-edited dictionaries. Throughout this work, we trained and tested on isolated sentences. We removed repeated identical sentences in Subtitle. We reserved 5\% of the data as a dev set and 5\% as a test set. After filtering, our C4 training set had 87.5\,B words and Subtitle had 528\,M words.

\subsection{Sentence classification}
Similar to past work \cite{vertanen_imagination,adhikary_abbreviated}, we wanted to focus our models on sentences similar in style to the text AAC users may want to write. We did this by training a three-way classifier on top of DeBERTaV3 \cite{he_debertav3}. For the out-of-domain class, we used a newswire corpus \cite{chelba_billion} which we denote as \textit{News}. For the written in-domain class, we used Forum. For the spoken in-domain class, we used a combination of Daily and BOLT. 

Daily was the smallest with 46\,K sentences. We used NLPAug \cite{ma_nlpaug} to add 232\,K sentences, changing one or two words (selected at random) in the original sentences. Words were changed using a distilled version of RoBERTa \cite{liu_roberta}. This brought Daily to the same size (279\,K sentences) as BOLT. We took the first 558\,K sentences from Forum and News to create an equal amount of data for each of our three classes (in-domain written, in-domain spoken, and out-of-domain). 
During classification, we lowercased sentences and removed characters aside from A--Z, apostrophe, and space. We optimized hyperparameters as described in \cref{appendix_hyper_sent_classifier}. We used early stopping during training. Our final model had a classification accuracy of 85.7\% on unseen sentences. 

\subsection{Domain adaptation procedure}\label{domain-adaptation}
We scored each sentence in C4 and Subtitle using our classifier. To adapt the opt-350m LLM, we selected sentences with a written or spoken probability that met a threshold. To efficiently determine which threshold to use, we trained 12-gram models from scratch across a broad range of thresholds and found we obtained the majority of perplexity gains with a threshold of 0.90 for C4 and 0.75 for Subtitle. Using these thresholds, our adaptation datasets consisted of 4.0\,B words of C4 and 130\,M words of Subtitle. With these thresholds as a starting point, we then trained opt-350m models with a narrower range of thresholds centered on 0.90 for C4 and 0.75 for Subtitle. The results, detailed in \cref{appendix_failed_improvements}, confirmed the thresholds we chose.  

We first tested fine-tuning all the model's parameters on a combined adaptation set consisting of C4 (threshold 0.90), Subtitle (threshold 0.75), BOLT, and Daily. As shown in \cref{table_fine_tuning}, this reduced the perplexity on our dev sets from 2.30 to 2.19. Next, we explored a curriculum adaptation process consisting of a sequence of separate fine-tuning runs. Each run was on one of our five datasets: C4, Subtitle, Daily, BOLT, or Forum. Before each fine-tuning, we optimized hyperparameters using the model from the previous step. 

Given the sizes of our five datasets, we opted to first always fine-tune on C4 followed by Subtitle. We tested all possible orders of further fine-tuning on Daily, BOLT, and Forum. Fine-tuning on Forum did not help in any order, and thus we excluded it from the curriculum. We found fine-tuning on C4, Subtitle, BOLT, and then Daily was the best order, improving perplexity to 2.10. Combining BOLT and Daily in a single final step did slightly better at 2.09. However, simply fine-tuning using the combined BOLT and Daily data did slightly better still at 2.07. While the classified C4 and Subtitle datasets did not prove useful for improving the subword LLM, they will be useful for creating or improving other models (see \cref{section_other_models}). 

\begin{table}[tb]
\setlength{\tabcolsep}{1.9pt}
\centering
\small
\begin{tabular}{ l r r r r r }
\toprule
Adaptation data         & \multicolumn{2}{c}{Fine-tuning}        & \multicolumn{3}{c}{Perplexity} \\
\cmidrule(lr){2-3}
\cmidrule(lr){4-6}
                                 & type & steps & Written       & Spoken        & ~~Avg \\
\midrule 
---                              & ---    & --- & 2.41	& 2.19	& 2.30 \\
C4+Sub+BOLT+Daily                & full & 1     & 2.30	& 2.08	& 2.19 \\
C4$\vsa$Sub$\vsa$BOLT$\vsa$Daily & full & 4     & 2.16	& \textbf{2.03}	& 2.10 \\
C4$\vsa$Sub$\vsa$BOLT+Daily      & full & 3     & 2.14	& \textbf{2.03}	& 2.09 \\
BOLT+Daily                       & full & 1     & \textbf{2.11}	& 2.04	& \textbf{2.07} \\
\midrule
C4$\vsa$Sub$\vsa$BOLT+Daily      & LoRA & 3     & 2.17	& 2.06	& 2.11 \\
BOLT+Daily                       & LoRA & 1     & 2.13	& 2.05	& 2.09 \\
\bottomrule
\end{tabular}
\caption{Perplexity on the two dev sets using various adaptation approaches on opt-350m. Subtitle (denoted Sub) used a 0.75 threshold and C4 used a 0.90 threshold. $\vsa$ denotes separate fine-tuning steps. + denotes datasets merged in a single fine-tuning step. Best result in bold.
\label{table_fine_tuning}}
\end{table}

Additionally, we fine-tuned opt-350m using Low-Rank Adaptation (LoRA)~\cite{hu_lora}, which drastically reduces the number of trainable parameters for downstream tasks. As shown in \cref{table_fine_tuning} (bottom), LoRA yields slightly higher perplexities compared to full parameter fine-tuning. As such, we continued to use full fine-tuning for the remainder of our experiments.

For additional results and hyperparameter values, see \cref{appendix_subword_hyper}.
We also explored other ways to try and improve adaptation. For details of things that did not work, see \cref{appendix_failed_improvements}.

\section{Other prediction approaches\label{section_other_models}}
With opt-350m, our search algorithm (\cref{sec_search_alg}) takes on average 79\,ms to predict the next character on an A100 GPU. While this may be fast enough for many purposes, interfaces such as Dasher~\cite{ward_dasher} or less capable devices may require more efficient prediction. We explored several other methods for obtaining faster predictions.

\subsection{N-gram model}
Inference with an n-gram model was very fast at 0.1\,ms (\cref{table_compare}). To serve as a baseline, and to provide an open training data n-gram model, we created a character 12-gram n-gram mixture model from C4 (threshold 0.90) and Subtitle (threshold 0.75) using SRILM \cite{stolcke_srilm}. We created separate C4 and Subtitle language models using Witten-Bell smoothing and no count cutoffs. We built a mixture model using linear interpolation with weights optimized to minimize perplexity on dev sentences from Forum, BOLT, and Daily. The  mixture weights were 0.84 (C4) and 0.16 (Subtitle).

The unpruned mixture model (denoted \textit{C4+Sub n-gram full}) was large with 5\,B parameters and a compressed size of 30\,GB. We used entropy pruning \cite{stolcke_entropy} to to reduce the model to 71\,M parameters and a compressed size of 488\,MB (denoted \textit{C4+Sub n-gram}). This was similar in size to the AAC n-gram we previously compared with (\cref{table_compare}). On our dev sets, our full and pruned models had perplexities of 2.53 and 2.60 respectively. Our pruned model's perplexity was somewhat higher than the 2.55 of the existing AAC n-gram model.


\subsection{Byte model}
The multilingual encoder-decoder ByT5 model \cite{xue_byt5} eschewed the use of subword tokenization and instead used a token set based on individual characters. ByGPT5 \cite{belouadi_bygpt5} is a decoder-only version of ByT5 that can make character predictions directly. 

The average perplexity on our dev sets of the base ByGPT5 model (289\,M parameters) was 2.85. We compared our two best adaptation methods we previously found on opt-350m (\cref{domain-adaptation}) using ByGPT5. First, we used a three step curriculum approach: adapting on C4, then on Subtitle, and finally on BOLT combined with Daily. This model had a perplexity of 2.14. 

Second, we fine-tuned the base model in a single-step on BOLT combined with Daily. This results in a higher perplexity of 2.27. We suspect due to the multilingual nature of the base model, first fine-tuning on C4 and Subtitle helped move the model towards English. For results and hyperparameter values at each phase of the fine-tuning, see \cref{appendix_byte_results}.


\begin{table*}[tb]
\setlength{\tabcolsep}{2.7pt}
\centering
\small
\begin{tabular}{ l l l r r r r r r r }
\toprule
Language  & Domain  & Character  & Written  & Spoken & AAC user & ALS bank & COMM2 & Average & Time  \\
 model &  adaptation &  prediction &  (ppl)  &  (ppl) &  (ppl) & (ppl) & (ppl) & (ppl) & (ms) \\
\midrule
AAC n-gram         & - & KenLM & 2.56	& 2.51	& 2.89	& 2.33	& 2.41	& 2.54	& \textbf{0.1} \\
C4+Sub n-gram      & - & KenLM & 2.62	& 2.54	& 2.86	& 2.43	& 2.45	& 2.58	& \textbf{0.1} \\
C4+Sub n-gram full & - & KenLM & 2.57	& 2.47	& 2.82	& 2.38	& 2.40	& 2.53	& \textbf{0.1} \\
\midrule
opt-350m           & -                                & search alg. & 2.40	& 2.15	& \textbf{2.24}	& 2.39	& 2.20	& 2.28	& 79.1 \\
opt-350m           & C4$\vsa$Sub$\vsa$BOLT$\vsa$Daily & search alg. & 2.14	& \textbf{2.00}	& 2.29	& 2.14	& \textbf{2.01}	& 2.12	& 78.9 \\
opt-350m           & C4$\vsa$Sub$\vsa$BOLT+Daily      & search alg. & 2.12	& \textbf{2.00}	& 2.30	& 2.14	& 2.02	& 2.12	& 78.2 \\
opt-350m           & BOLT+Daily                       & search alg. & \textbf{2.10}	& 2.01	& 2.28	& \textbf{2.13}	& \textbf{2.01}	& \textbf{2.11}	& 78.0 \\
\midrule
ByGPT5             & C4$\vsa$Sub$\vsa$BOLT+Daily      & directly    & 2.20	& 2.04	& 2.57	& 2.18	& 2.06	& 2.21	& 11.6 \\
opt-350m           & C4$\vsa$Sub$\vsa$BOLT+Daily      & classifier  & 2.42  & 2.42  & 3.03  & 2.38  & 2.36  & 2.52  & 14.6 \\
\bottomrule
\end{tabular}
\caption{Final evaluation of our various language models and methods for predicting the next character. $\vsa$ denotes separate fine-tuning steps. + denotes datasets merged in a single fine-tuning step. Inference times are the average of per-character inferences over the test sets. The best result in each column is shown in bold.
\label{table_final_eval_ppl}}
\end{table*}

\begin{table*}[tb]
\setlength{\tabcolsep}{2.7pt}
\centering
\small
\begin{tabular}{ l l r r r r r r }
\toprule
Language model & Domain adaptation    & Written & Spoken & AAC user & ALS bank & COMM2 & Average\\
\midrule
AAC n-gram         & -	& 59.7\%	& 60.0\%	& 55.3\%	& 62.2\%	& 61.0\%	& 59.7\%	\\
C4+Sub n-gram      & -	& 59.2\%	& 59.9\%	& 56.4\%	& 61.1\%	& 61.2\%	& 59.6\%	\\
C4+Sub n-gram full & -	& 60.1\%	& 61.4\%	& 57.4\%	& 61.8\%	& 62.0\%	& 60.5\%	\\
\midrule
opt-350m           & - 	& 61.7\%	& 65.1\%	& 63.1\%	& 61.9\%	& 64.2\%	& 63.2\%	\\
opt-350m           & C4$\vsa$Sub$\vsa$BOLT$\vsa$Daily	& 64.6\%	& 67.1\%	& 62.5\%	& 64.5\%	& \textbf{66.4}\%	& 65.0\%	\\
opt-350m           & C4$\vsa$Sub$\vsa$BOLT+Daily 	& 64.9\%	& 67.0\%	& 62.3\%	& 64.4\%	& 66.3\%	& 65.0\%	\\
opt-350m           & BOLT+Daily 	& \textbf{65.2}\%	& \textbf{67.2}\%	& \textbf{62.7}\%	& \textbf{64.7}\%	& 66.3\%	& \textbf{65.2}\%	\\
\midrule
ByGPT5 & C4$\vsa$Sub$\vsa$BOLT+Daily & 64.3\%	& 66.8\%	& 61.6\%	& 63.7\%	& 65.9\%	& 64.5\%	\\
\bottomrule
\end{tabular}
\caption{Keystroke savings for language models using an onscreen predictive keyboard with five word predictions. The best result in each column is shown in bold.
\label{table_final_eval_ks}}
\end{table*}

\subsection{Classification\label{classifier_section}}
We also explored making character predictions by adding a classification layer to the LLM. Instead of using the model to generate the ensuing text from some context, we asked it to produce a distribution over labels -- each label represented a character. We built the model on top of the best domain-adapted opt-350m model from \cref{domain-adaptation}. This retained the model's domain knowledge and focused additional training on the classification task. 

To create training examples, we selected a random location within each sentence. We truncated the sentence at that point and used the character that would come next as the label. If the next character was not in our symbol set (A--Z, a--z, space, and apostrophe), we chose another random location. We utilized the same three step curriculum approach as the byte model; we first trained on C4, then Subtitle, and finally on BOLT combined with Daily. For the smaller BOLT and Daily sets, we added a hyperparameter for the number training examples generated per sentence. During training, we updated the classification layer as well as all other model parameters.

The classifier model did not perform as well as we had hoped, achieving a perplexity of 2.47. By comparison, training with BOLT and then Daily in separate steps yielded a slightly higher perplexity of 2.54. Training on just BOLT and Daily performed worse at 2.60. See \cref{appendix_class_results} for additional results and hyperparameter values.


\section{Final evaluation}\label{sec_final_eval}

At this point, we have extensively used our private dev sets to guide our process. We conducted a final evaluation with our unseen private test sets and three additional test sets. Firstly, we created a test set based on the text of an AAC user. This set consisted of 3.6\,K sentences (65\,K words) written by a user of Dasher \cite{ward_dasher}. The text was from public and private presentations given by the user. We also added two public test sets: voice banked messages of people with ALS~\cite{costello_message}, and the COMM2 set of responses to conversational situations~\cite{vertanen_comm2}. 

As shown in \cref{table_final_eval_ppl} (top), our C4+Sub n-gram model performed slightly worse with an average test set perplexity of 2.58 compared to the AAC n-gram model which averaged 2.54. Our unpruned full model did slightly better at 2.53. All n-gram models were outperformed by the LLMs, both with and without domain adaptation.

As shown in \cref{table_final_eval_ppl} (middle), using the search algorithm with opt-350m adapted in a single step on the combination of BOLT and Daily performed the best on two of the five test sets and the overall average. It tied with the four-step curriculum on the COMM2 set (2.01). The base opt-350m yielded the lowest perplexity on the test set from the AAC user (2.24), though the curriculum adapted opt-350m models were not far behind (2.28--2.30). 

Averaging the perplexity across all models, there was a noticeable increase in perplexity between our written (2.35) and spoken (2.24) test sets and the presentations given by the AAC user (2.59). The public test sets also produced lower average perplexities (2.28 for ALS bank and 2.21 for COMM2). While we hoped our data selection approach using diverse text from mobile forum messages would allow our adapted models to perform well on both simple everyday communications and more complex planned written communications, at least for this one user, this did not work. As a measure of text complexity, we calculated the number of characters per sentence. The AAC user's sentences were on average 94.4 characters per sentence compared to the much shorter sentences in the written (31.4) and spoken (39.9) test sets.

Of the LLM approaches, the byte LLM provided the fastest predictions at around 12\,ms. This was 6.8 times faster than our search algorithm and 1.3 times faster than using a classification layer. 

To demonstrate the potential impact in a real-world AAC interface, we simulated using each language model in a onscreen keyboard with five word predictions. We simulated the keystrokes a hypothetical perfect user would save compared to typing every character. \textit{Keystroke savings} is calculated as:
\begin{equation*}
\frac{k_a - k_p}{k_a}\times100\%,
\end{equation*}
where $k_{a}$ is the number of keystrokes required without word predictions and $k_p$ is the number of keystrokes required with predictions. Higher keystroke savings is better. 

The keyboard was allowed to make up to five word predictions. Predictions were made before any characters were typed for the next target word as well as during each subsequent prefix of the target word. The simulated user made no typing errors. We did not filter the offered word predictions based on whether the user did not choose a given prediction for a shorter prefix for the current target. If the target word appeared in the predictions, we assumed one key press selected the target word and added any following space.

As shown in \cref{table_final_eval_ks}, keystroke savings improvements followed the perplexity gains (\cref{table_final_eval_ppl}). Compared to the AAC n-gram, across the test sets our best LLM provided a 5.5\% absolute increase in keystroke savings. To contextualize this, imagine someone is using an eye-tracker and an onscreen keyboard with a dwell time of one second. Assuming they make no errors and incur no other overheads, writing the average length sentence in our spoken test set would take about 40 seconds  without word predictions. Making optimal use of predictions provided by the AAC n-gram would take about 16.0 seconds. Using our best model (opt-350m adapted on BOLT+Daily), the user would take about 13.1 seconds, 18\% faster than using the n-gram model for predictions.

\section{Discussion\label{sec_discussion}}

Most of our trained models had higher perplexities on the AAC test set than on any of the other test sets. The only exception was the base opt-350m model, which performed worst on the written test set. We believe this stems from the nature of the test sets. The written and spoken sets were created by crowdsourced workers writing about topics such as the weather or travel. As described in \cref{sec:appendix_a}, workers were instructed to invent questions or statements they might include in a conversation with another person. The ALS bank and COMM2 sets contain text of a similar nature. This aligns closely with the in-domain conversational datasets we used in our adaptation process. 

The AAC test set was derived from prepared presentations by an AAC user with ALS discussing their life experiences. It covers more complex topics and uses more advanced language. Presentations are also not conversational in nature, which is why we believe our curriculum adaptation process was not effective at improving the model's performance on the AAC test set. We also note that AAC users are diverse in nature and our domain adaptation may be advantageous to other users or even to this user in other communication situations. It may also be advantageous to consider adapting the language model more directly to a particular user in lieu of, or in addition to, domain adaptation. 

We obtained the lowest perplexities using a subword tokenized LLM and our search algorithm. While we investigated a multi-step curriculum to adapt the opt-350m model using sentences classified from large out-of-domain datasets, a simple fine-tuning using in-domain datasets worked better. The main drawback of the search algorithm is its inference time. If faster predictions are required for a particular interface or device, a byte or classification model may be preferred. However, currently subword LLMs are much more popular than byte LLMs. Further, the search approach can be used directly on an LLM without task-specific training.


Adding a classification layer to the domain-adapted opt-350m model performed worse than any of the other LLM-based approaches, including the base opt-350m model before domain adaptation. We have a few theories as to why this may have been the case. Our classification training examples chose a random location in each sentence, so many of the sentences included in these further training steps were likely fragmented, which may have affected earlier layers in the model. To investigate this theory, we could freeze the lower model layers during the classification training. This might help to preserve the domain adaptation and prior model while also tuning the new classification layer. 

It is also possible that the subword nature of the model prevented it from fully learning the character classification task. As discussed in \cref{sec_search_alg}, adding a new character can rearrange a word's optimal tokenization. Because the model was originally trained to predict the next token in a sequence, it may not have considered all the different possible tokenizations as our search algorithm does.

\section{Conclusions\label{sec_conclusions}}

We investigated how to leverage large transformer language models to aid non-speaking individuals using Augmentative and Alternative Communication (AAC) devices. In particular, we focused on how to generate the character predictions needed by interfaces where users write one letter-at-a-time. We presented an algorithm for obtaining character probabilities from LLMs using subword tokenization, and compared our algorithm to using a byte-level LLM or adding a classification layer on top of a subword LLM. We detail how we classified sentences in large corpora of web and movies subtitles text based on how similar sentences were to written and spoken communications. We investigated a multi-step fine-tuning process using our scored sentences as well as our smaller in-domain corpora to adapt LLMs to our target domain.

We found that we were able to produce the lowest per-character perplexities using our search algorithm with the subword opt-350m model. We found that our domain adaptation curriculum was effective at improving the model performance for simple, conversational text, but did not improve performance for more complex prepared presentations from a single AAC user. Though it offered the best perplexity, our subword to character algorithm took almost seven times longer to produce character predictions than a native byte-level LLM. 

To aid future research, we have made our unique datasets and models available.\footnote{\url{https://osf.io/ajm7t/}} This includes the model for classifying sentences as written or spoken, scored C4 and Subtitle sentences, n-gram models trained on the classified sentences, fine-tuned opt-350m model, fine-tuned ByGPT5 model, and letter classification model.

\section*{Limitations}

The ultimate goal of this work was to adapt large language models for use in AAC devices. With that goal in mind, it is a limitation that we did not have training data from actual AAC users. As we discussed in \cref{private_test_data}, such data is difficult to come by, so the datasets we created and used in this work aimed to approximate AAC-like text. While we think our models can provide an improved initial experience for AAC text entry interfaces, we think much more substantial improvements are likely possible if models are adapted using the actual text of a given user. Doing this in a practical and privacy-preserving way will need further research.

The experiments we conducted as part of this research consisted solely of the offline evaluation of our models with fixed datasets. While the perplexity values and keystroke savings we report are generally considered to be indicative of a model's performance, they may not show the full extent of the impact on an actual user entering text. Text input is a complex task, especially when performed through an AAC device. Further research and user evaluation is required to fully understand the benefit or detriment of our tuned language models on the text entry performance of AAC users.

Some recent work has explored freezing model layers that have relatively low training loss in order to focus fine-tuning on improving underperforming layers~\cite{yusufali_refining_2024}. This was not a technique that we were able to explore in the scope of this work. However, we may have been able to improve the performance of the classification model by freezing the lower model layers that were already domain adapted and focusing the additional training on the classification layer specifically. 


We only tested our search algorithm with a few model architectures. Of particular concern is our handling of token removal. Because the tokenizer used by the models in this work includes a space at the beginning of many of its tokens, our algorithm removes the end of the context up to and including the last space. Other models may have tokenizers that handle space differently. Our algorithm may need adjustment to work with such models.

The main advantage of our search algorithm compared to the byte model is that it works out of the box with the most common LLMs which use subword tokenization. However, the main disadvantage is obtaining the distribution over the next character takes substantially longer. The search itself in many cases needs to make several inferences on the GPU so we can expect it to take several times as long as a byte LLM or classification head-based approach which can make the character prediction in a single inference. During the development of our search algorithm, we conducted multiple rounds of performance optimization. We introduced various changes to the algorithm that, in the end, resulted in performance that was several orders of magnitude faster than our initial prototype. However, it is possible additional efforts could further improve performance. 

Currently, our search algorithm is optimized for inference on a GPU and not a CPU. This could limit the AAC devices our algorithm can practically be deployed on. However, for many AAC interfaces, such as those based on switch or brain input, waiting a tenth of a second for a prediction is likely acceptable and would not unduly impact user experience.

We primarily focused on fine-tuning approaches to domain adaptation and the task of classifying the next character. In-context learning (ICL) has emerged in recent years as an alternative for fine-tuning~\cite{dong_in_context}. With in-context learning, a few examples of the task or domain are provided to the language model along with the query. The model takes the examples into account when producing its response, instead of requiring a separate training procedure. While ICL has shown promise, such emergent abilities may require large model sizes~\cite{wei_emergent}. For privacy reasons, our focus here was on modest sized models that could be plausibly used and fine-tuned on an end user's device. We were not able to explore the feasibility of ICL using modest sized models in the scope of this paper, and leave this as future work. 

\section*{Ethical considerations}
Our fine-tuning process presented in this work relied heavily upon the Forum~\cite{vertanen_mining}, BOLT~\cite{song_bolt}, and DailyDialog~\cite{li_dailydialog} datasets. While we did introduce the additional larger datasets C4~\cite{raffel_t5} and Subtitle~\cite{lison_subtitles}, we filtered those sets to sentences that were similar to our in-domain sets. This achieved our intended result of improving the performance of our character prediction models. However, any biases present in Forum, BOLT, and Daily may also be reflected in the text selected from C4 and Subtitle. This could lead to predictions made by text entry interfaces relying on our models to skew towards those biases.

When developing AAC interfaces that leverage language models to accelerate user input, it is crucial to support user autonomy, i.e.~allowing users to express exactly what they want and not just what is probable under a language model. We developed our models to assist users in entering text based on the text that is most likely. However, there may be cases where users wish to enter text that is not likely, or that the model has not seen before. We encourage anyone who uses our models to design their interfaces such that users can control to what degree, if any, the language model contributes to the produced text.

\section*{Acknowledgments}
We thank the AAC user who provided their text for use in our final evaluation. We thank Soufia Bahmani for her work on the subword LLM word prediction algorithm used in \cref{table_final_eval_ks}. This work was funded by the National Institutes of Health / National Institute on Deafness and Other Communication Disorders (R01DC009834) and by the National Science Foundation (IIS-1750193 and IIS-2402876). Any opinions presented in this work are those of the authors and do not reflect the opinions of our funding agencies. 

\bibliography{references,emnlp2023}

\appendix

\section{Private test set details\label{appendix_test_set}}

\subsection{Written test set}
We obtained compositions by crowdsourced workers collected by \citet{vertanen_complementing}. With the exception of a table of examples and the text from the AAC condition in Experiment 1, these novel compositions were never publicly available. Workers were asked to imagine they were writing on a mobile device and that they should invent a ``fictitious but plausible message''. We used the data from condition \textsc{Compose} in Experiment 1, both conditions in Experiment 2, and the \textsc{Compose} condition in Experiment 3. See \citet{vertanen_complementing} for further details. 

We reviewed workers' text using a semi-automated process. We expanded abbreviations, corrected spelling and grammar errors, and split compositions into sentences. If we could not determine a worker's intended text, we dropped the composition. \cref{table_written_examples} shows some examples from the test set. We subsequently removed these sentences as well as those appearing in tables in \citet{vertanen_complementing} from the test set. We also removed any sentences containing potentially identifiable information.

\begin{table}[tb]
\setlength{\tabcolsep}{4pt}
\centering
\small 
\begin{tabular}{l}
\toprule
Are you going to the party? \\
Do you want to go out for dinner tonight? \\
Wanna go to taco bell? \\
I am stuck in a meeting but will call you when I get out. \\
is it just texting that's outlawed while driving or talking too \\
Be sure to watch the meteor shower tonight! \\
I lost the book. \\
sarah never called me back \\
\bottomrule
\end{tabular}
\caption{Examples of written style communications.\label{table_written_examples}}
\end{table}

We took only the unique compositions from a given worker. We only used workers who rated their English ability as native. We assigned each unique worker an anonymous number. A total of 227 unique workers were in our dataset. We made a dev and test set from the even and odd numbered participants respectively.  For the purposes of the experiments conducted here, we lowercased text and stripped end-of-sentence punctuation, dashes, and commas. We dropped sentences containing other characters (e.g.~numbers). Our dev and test sets contained 1,348 lines (8,645 words) and 1,347 lines (9,044 words) respectively.

\subsection{Spoken test set}
Similar to past work \cite{venkatagiri_layouts,vertanen_comm2}, we asked people to each write eight different questions or statements in response to an everyday communication situation (see \cref{sec:appendix_a} for our exact task wording). We did this on Amazon Mechanical Turk. Each worker received one of the following situations: 
\begin{itemize}[left=0mm,labelsep=1mm,parsep=0mm]
    \item \textbf{Opening} -- Starting a conversation with someone. 
    \item \textbf{Weather} -- Talking about the weather.
    \item \textbf{Meals} -- Talking about eating and meals.
    \item \textbf{Travel} -- Talking about trips and traveling.
    \item \textbf{Closing} -- Ending a conversation with someone. 
\end{itemize}

Our study was reviewed by our institutional review board (IRB) and judged to be exempt. Workers were paid \$0.80 USD to complete the study. First, workers completed a consent form and then created eight unique sentences. While we asked workers to rate their English ability, we found this rating was often not accurate. We reviewed each worker's sentences and eliminated 54 workers who had poor English or misunderstood the task (e.g.~talking about strategies for ending a conversation rather than what you might actually say). This left us with 246 unique workers who were each assigned an anonymous number.

Occasionally, we found workers wrote sentences that were conversational in nature, but did not match the target situation. We suspect this was due to worker inattention. We left these sentence in our test set since we felt they were representative of conversational messages an AAC user might want to write. We semi-automatically corrected obvious mistakes in spelling and grammar, and split compositions comprised of multiple sentences. We removed any sentences containing potentially identifiable information.

\cref{table_spoken_examples} shows examples of sentences written in each situation. In general, we found workers' compositions were creative and relevant to the target situation. After removing the sentences in \cref{table_spoken_examples}, we made a dev and test set from the even and odd numbered participants respectively. For the purposes of the experiments conducted here, we lowercased text and stripped end-of-sentence punctuation, dashes, and commas. We dropped sentences containing other characters (e.g.~numbers). Our dev and test sets contained 1,469 lines (11,261 words) and 1,360 lines (10,821 words) respectively. 

\begin{table}[tb]
\setlength{\tabcolsep}{4pt}
\centering
\small 
\begin{tabular}{l}
\toprule
\textbf{\textit{Opening:}}\\
\hspace{2mm} How old are you? \\
\hspace{2mm} Good morning Andrea. \\
\hspace{2mm} Sorry, Paul i have lots of work in my office.\\
\midrule
\textbf{\textit{Weather:}}\\
\hspace{2mm} The roads will be hard to see because of the fog. \\
\hspace{2mm} How many people were killed during Katrina? \\
\hspace{2mm} Is it going to rain next weekend? \\
\midrule
\textbf{\textit{Meal:}}\\
\hspace{2mm} Do you prefer chicken or beef? \\
\hspace{2mm} What's your favorite kind of meat?\\
\hspace{2mm} Isn't this good? \\
\midrule
\textbf{\textit{Travel:}}\\
\hspace{2mm} What are you looking forward to seeing? \\
\hspace{2mm} What do you think the food scene is like in Europe? \\
\hspace{2mm} What is the best place for a vacation in your country? \\
\midrule
\textbf{\textit{Closing:}}\\
\hspace{2mm} I'm sorry, but I have to go. \\
\hspace{2mm} It was a pleasure catching up with you.\\
\hspace{2mm} I will see you around.\\
\bottomrule
\end{tabular}
\caption{Examples of spoken style communications in our five situations.\label{table_spoken_examples}}
\end{table}

\subsection{Genre comparison}
We found sentences in the written set were shorter than in the spoken set (6.6 versus 7.8 words per sentence). We suspect this was due to the mobile input scenario workers were asked to imagine in \citet{vertanen_complementing}. However, given the input rate limits faced by many AAC users, the concise nature of the written set may be well-matched to our target users. 

For the spoken set, workers were free to choose between writing a statement or a question. We classify sentences as a question, statement, or exclamation based on their end-of-sentence punctuation (i.e.~period, question mark, or exclamation point). In our spoken set, 60\% were questions, 36\% were statements, and 4\% were exclamations. In the written set, 25\% were questions, 43\% were statements, and 11\% were exclamations.

\subsection{Crowdsourced task}

\label{sec:appendix_a}
\begin{figure}[tb]
  \includegraphics[width=1.0\linewidth]{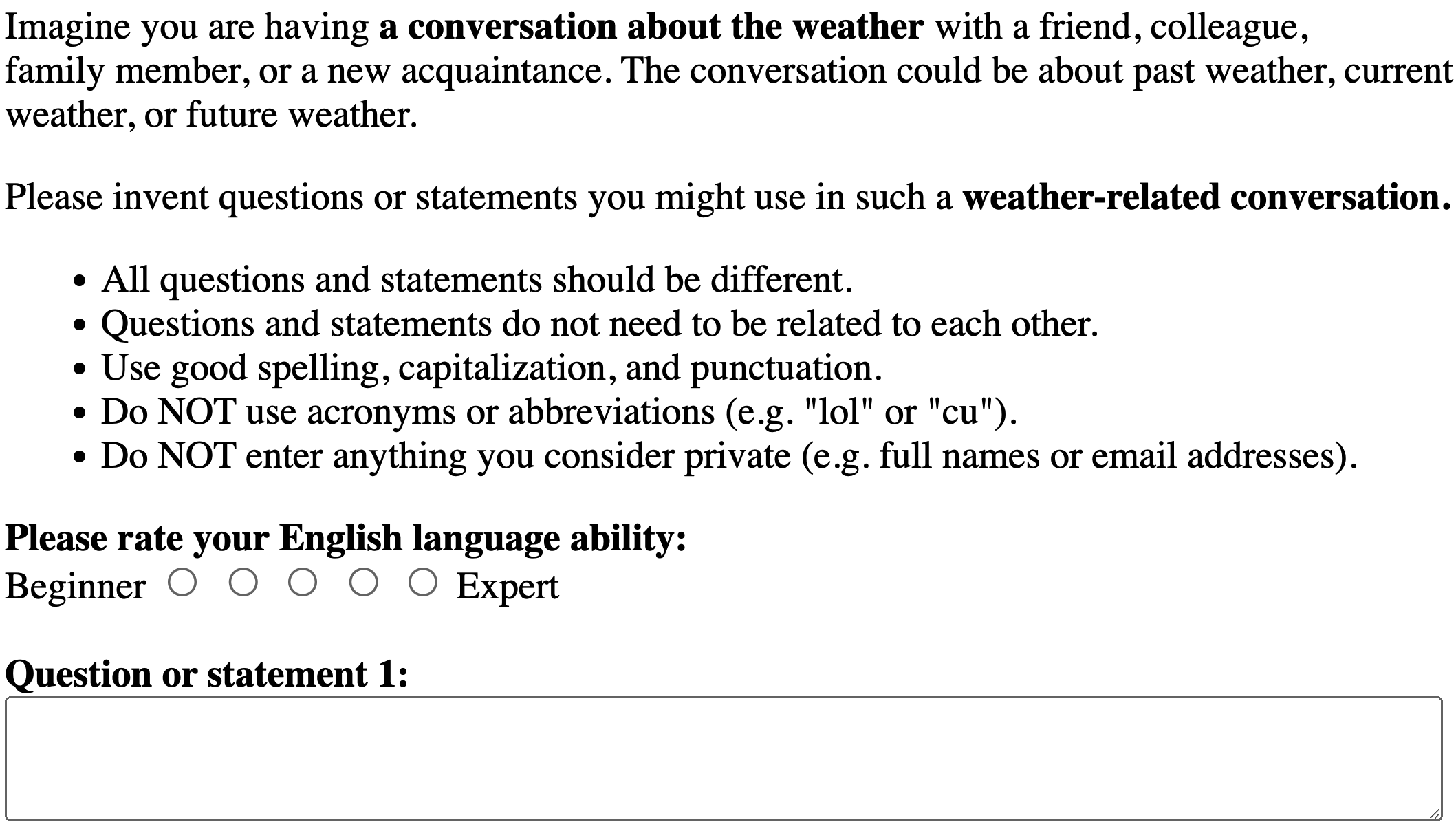}\\
  \caption{Worker instructions for our conversational text collection task (weather communication situation).}
  \label{figure_hit}
\end{figure}

\cref{figure_hit} shows the web interface we used to collect AAC-like communications from workers on Amazon Mechanical Turk. We changed the text at the top based on the situation as follows:
\begin{itemize}[left=0mm,labelsep=1mm,parsep=0mm]
    \item \textbf{Opening} -- ``Imagine you are \textbf{starting a conversation} with a friend, colleague, family member, or a new acquaintance. Please invent questions or statements you might use at the \textbf{start of the conversation}.''
    \item \textbf{Weather} -- ``Imagine you are having \textbf{a conversation about the weather} with a friend, colleague, family member, or a new acquaintance. The conversation could be about past weather, current weather, or future weather. Please invent questions or statements you might use in such a \textbf{weather-related conversation}.''
    \item \textbf{Meals} -- ``Imagine you are having \textbf{a conversation about meals} with a friend, colleague, family member, or a new acquaintance. The conversation could be about past meals, current meals, or future meals. Please invent questions or statements you might use in such a \textbf{meal-related conversation}.''
    \item \textbf{Travel} -- ``Imagine you are having \textbf{a conversation about traveling} with a friend, colleague, family member, or a new acquaintance. The conversation could be about past travels, current travels, or future travels. Please invent questions or statements you might use in such a \textbf{travel-related conversation}.''
    \item \textbf{Closing} -- ``Imagine you are speaking with a friend, colleague, family member, or a new acquaintance. You have reached the point where \textbf{you would like to end the conversation}. Please invent questions or statements you might use use to \textbf{signal you would like to end the conversation}.'' 
\end{itemize}

\section{Additional results and details}

\subsection{Sentence classifier hyperparameters\label{appendix_hyper_sent_classifier}}
We created a model to classify sentences from C4 and Subtitle on top of DeBERTaV3 \cite{he_debertav3}. We used a linear learning rate scheduler. We optimized the model's hyperparameters via grid-search as in Table 10 of \citet{he_debertav3}. 

We searched over batch sizes of [16, 32, 48, 64], learning rates of [1.5e-5, 2.0e-5, 2.5e-5, 3.0e-5], warmup steps of [50, 100, 500, 1000], and dropout of [0.0, 0.1, 0.15]. Weight decay was fixed at 0.01. Hyperparameter settings were evaluated based on held out sentences from the the Forum, BOLT, Daily, and News datasets. Our optimal hyperparameters were: batch size 48, learning rate 2.5e-05, warmup steps 100, and dropout 0.0. 

\begin{table*}[tb]
\setlength{\tabcolsep}{4pt}
\centering
\small
\begin{tabular}{ l l r r r r r r r r r r  }
\toprule
 Adaptation data & \multicolumn{2}{c}{Fine-tuning} & Dropout & Warmup & Stable & Learning & Weight & Batch & Epochs & Tuner & Avg \\
\cmidrule(lr){2-3}
 & type & steps & & steps & & rate & decay & size &  & trials & (ppl) \\
\midrule 
opt-350m base & --- & --- & --- & --- & --- & --- & --- & --- & --- & --- & 2.30 \\
\midrule
~~$\vsa$~C4+Sub+BOLT+Daily                   & full & 1 & .698 & 751 & .586 & 1.77e-06 & .468 & 1120 & 1  & 88   & 2.19 \\    
\midrule
~~$\vsa$~C4                                  & full & 1 & .142 & 668 & .903 & 2.07e-06 & .830 & 2240 & 1  & 80   & 2.22 \\ 
~~~~~$\vsa$~Sub             & full & 2 & .245 & 546 & .597 & 6.71e-08 & .025 & 7840 & 1  & 114  & 2.17 \\  
~~~~~~~~$\vsa$~BOLT        & full & 3 & .178 & 590 & .830 & 2.99e-07 & .070 & 6720 & 10 & 849  & 2.11 \\    
~~~~~~~~~~~$\vsa$~Daily   & full & 4 & .435 & 900 & .175 & 3.53e-04 & .288 & 5600 & 1  & 4252 & 2.10 \\
~~~~~~~~$\vsa$~BOLT+Daily  & full & 3 & .650 & 49 & .638 & 1.61e-06 & .547 & 2240 & 1  &  379  & 2.09 \\    
\midrule
~~$\vsa$~BOLT+Daily                          & full & 1 & .483 & 905 & .009 & 1.73e-05 & .323 & 6720 & 9 &  267  & 2.07 \\    
\midrule
~~$\vsa$~C4                                  & LoRA & 1 & .111 &   6 & .988 & 1.59e-04 & .389 & 7840 & 1 &   75  & 2.21  \\    
~~~~~$\vsa$~Sub             & LoRA & 2 & .003 & 442 & .587 & 1.39e-06 & .368 & 6720 & 1 &  101  & 2.18 \\    
~~~~~~~~$\vsa$~BOLT+Daily  & LoRA & 3 & .385 &  18 & .040 & 5.20e-05 & .535 & 7840 & 1 & 387 &  2.12 \\    
\midrule
~~$\vsa$~BOLT+Daily  & LoRA & 1 & .524 &  98 &   .983 & 5.73e-04 & .898 & 7840 & 14 & 92 &   2.09 \\    
\bottomrule
\end{tabular}
\caption{Hyperparameters after the specified number of tuning trials during our adaptation process on the opt-350m LLM. Also shown is the average perplexity on the written and spoken dev sets. We used a sentence selection threshold of 0.90 for C4 and 0.75 for Subtitle. $\vsa$ denotes separate fine-tuning steps. + denotes datasets merged in a single fine-tuning step. \label{table_hyper_subword}}
\end{table*}
\subsection{Subword hyperparameter tuning\label{appendix_subword_hyper}}
We arrived at our final subword opt-350m model by first fine-tuning on C4 (threshold 0.90), then Subtitle (threshold 0.75), then BOLT, and finally Daily. At each step, we tuned hyperparameters using Optuna \cite{optuna_2019} for the equivalent of 2 GPU days on an A100 GPU. 

We used a warmup stable decay (WSD) learning rate scheduler \cite{hu_wsd}. We optimized with respect to loss on held out sentences in Daily, BOLT, and Forum. The held out sentences were in a 1:1:2 proportion for Daily, BOLT, and Forum respectively to balance the influence of the written and spoken text genres. We used an 8-bit version of the AdamW optimizer. 

Fine-tuning examples were single sentences in mixed case with punctuation and started with the model's default start token of \texttt{</s>}. For C4 and Subtitle, we tuned hyperparameters using four million sentences chosen at random from the C4 0.90 and Subtitle 0.75 datasets to allow a more thorough search of the parameter space in the allotted time. Once we selected hyperparameters, we trained the model using the entire dataset. For Daily and BOLT we used all the training data from the datasets for both hyperparameter tuning and training. 


We tuned the following hyperparameters, allowing each to vary within the given ranges:
\begin{itemize}[left=0mm,labelsep=1mm,parsep=0mm]
    \item \textbf{Dropout} --- Dropout rate during fine-tuning, [0.0, 0.75].
    \item \textbf{Warmup steps} --- Number of warmup steps in the WSD learning rate scheduler, [0, 1000]. 
    \item \textbf{Stable} --- Proportion of the non-warmup steps in the WSD scheduler that were at a stable (fixed) learning rate, [0.0, 1.0]. 
    \item \textbf{Learning rate} --- Learning rate for the stable phase of the WSD scheduler, [1e-9, 1e-3].  
    \item \textbf{Weight decay} --- Weight decay during training, [1e-4, 1.0].  
    \item \textbf{Batch size} --- Training batch size, 1120, 2240, 3360, 4480, 5600, 6720, 7840, or 8960. The batch size was chosen to maximize the memory utilization of an A100 GPU with 40\,GB of memory and assuming final model training was done using distributed data parallel (DDP) on four GPUs. 
    \item \textbf{Epochs} --- Training epochs over a dataset. We fixed the training epochs to 1 for C4 and Subtitle. We allowed the training epochs to be in [1, 20] for BOLT and Daily. 
\end{itemize}

\cref{table_hyper_subword} shows the tuned hyperparameter values as well as the average dev set perplexity for each step of our adaptation of opt-350m. Each step of the fine-tuning curriculum lowered the perplexity on the dev sets. However, we found simply tuning in one-step on a combination of BOLT and Daily yielded the best results, reducing the perplexity of the original opt-350m model from 2.30 to 2.07 (a 10\% relative improvement). 

We compared the utility of each of our datasets in isolation in \cref{table_isolated_datasets}. We found BOLT performed the best, followed by Daily, Subtitle 0.75, C4 0.90, and finally Forum. To show that selecting sentences using our classification model helped, we fine-tuned on an equivalent random amount of C4 and Subtitle text. As shown in \cref{table_isolated_datasets} (bottom), random data performed worse than using the threshold selected data. In the case of C4, random data degraded predictions compared to the base model.

\begin{table}[tb]
\setlength{\tabcolsep}{4pt}
\centering
\small
\begin{tabular}{ l r r r }
\toprule
                    & Written & Spoken & Avg \\
                    & (ppl)   & (ppl)  & (ppl) \\
\midrule
opt-350m base        & 2.41	& 2.19      & 2.30  \\
\midrule
~~$\vsa$~Forum             & 2.37    & 2.19      & 2.28  \\ 
~~$\vsa$~C4 0.90           & 2.34    & 2.09      & 2.22  \\ 
~~$\vsa$~Subtitle 0.75     & 2.25    & 2.13      & 2.19  \\ 
~~$\vsa$~Daily             & 2.25    & 2.07      & 2.16  \\ 
~~$\vsa$~BOLT              & 2.19    & 2.09      & 2.14  \\ 
\midrule
~~$\vsa$~C4 random         & 2.47    & 2.17      & 2.32  \\ 
~~$\vsa$~Subtitle random   & 2.29    & 2.15      & 2.22  \\ 
\bottomrule
\end{tabular}
\caption{Perplexity of opt-350m on the dev sets after single-step fine-tuning on different datasets. $\vsa$ denotes separate fine-tuning steps.\label{table_isolated_datasets}}
\end{table}

\subsection{Potential subword improvements\label{appendix_failed_improvements}}

\begin{table}[tb]
\setlength{\tabcolsep}{4pt}
\centering
\small
\begin{tabular}{ l r r r }
\toprule
                    & Written & Spoken & Avg \\
                    & (ppl)   & (ppl)  & (ppl) \\
\midrule
opt-350m base        & 2.41	& 2.19      & 2.30  \\
\midrule
~~$\vsa$~C4 0.95           & 2.36	& 2.12	     & 2.24  \\
~~$\vsa$~C4 0.90           & 2.34    & 2.09      & 2.22  \\
~~$\vsa$~C4 0.85           & 2.36	& 2.11	     & 2.23  \\
\midrule
~~$\vsa$~C4 0.90~$\vsa$~Subtitle 0.80 & 2.24 & 2.09 & 2.17  \\
~~$\vsa$~C4 0.90~$\vsa$~Subtitle 0.75 & 2.24 & 2.09 & 2.17  \\
~~$\vsa$~C4 0.90~$\vsa$~Subtitle 0.70 & 2.25 & 2.10 & 2.17  \\
\bottomrule
\end{tabular}
\caption{Perplexity of opt-350m on the dev sets using different probability thresholds for selecting sentences from the C4 and Subtitle datasets. $\vsa$ denotes separate fine-tuning steps.\label{table_threshold}}
\end{table}

During our experiments adapting the opt-350m subword LLM, we explored various things that did not improve the perplexity on our dev sets. 

Recall that we used n-gram models trained on data selected from C4 and Subtitle at different thresholds. We found a threshold of 0.90 for C4 and a threshold of 0.75 for Subtitle provided the majority of perplexity gains when training an 12-gram language model from scratch. 

As shown in \cref{table_threshold}, increasing or decreasing the threshold by 0.05 on C4 increased perplexity. Subsequent adaptation of the C4 0.90 fine-tuned model on Subtitle performed best (or the same) with a threshold of 0.75 versus 0.70 or 0.80.

 
With recurrent neural network language models (RNNLMs), interpolating an RNNLM with an n-gram model can perform better than either model in isolation \cite{mikolov_learning}. We computed the per-character probabilities for our spoken and written dev sets using 1) the n-gram optimized for AAC-like text, 2) the original opt-350m LLM, and 3) our best fine-tuned opt-350m LLM. Using SRILM \cite{stolcke_srilm}, we found the optimal mixture weights between the n-gram and each LLM (the unadapted and fine-tuned opt-350m models). We found the optimal mixture weights with respect to the spoken and written dev sets independently.

For the unadapted LLM, the optimal mixture had a perplexity of 2.28, slightly lower than the 2.30 obtained using only opt-350m. For the spoken and written dev sets, the n-gram model received a mixture weight of 0.12 and 0.29 respectively. However, the fine-tuned LLM and n-gram mixture model had the same perplexity of 2.10 as using only the fine-tuned LLM. For the spoken and written dev sets, the n-gram model received a mixture weight of 0.04 and 0.03 respectively. Thus it seems the gains offered by the n-gram model's in-domain training data could be obtained instead by fine-tuning a pretrained LLM on in-domain data. 

We tried starting each adaptation example with a unique three-token sequence generated from the text ``\texttt{~AAC|}'' rather than the model's default start token of ``\texttt{</s>}''. We hoped this would allow the model to associate this token sequence with the AAC-like text seen during fine-tuning. This might encourage the model to generate more AAC-like text when seeing this sequence again at inference time. We compared the opt-350m model fine-tuned in a single step on the combined BOLT and Daily data. We tuned separate model hyperparameters for each start token sequence. Using a starting context of ``\texttt{~AAC|}'' resulted in an average perplexity on our dev sets of 2.21. Using the default starting context of ``\texttt{</s>}'' resulted in a slightly lower average perplexity of 2.18.

\subsection{Byte model adaptation\label{appendix_byte_results}}

\begin{table*}[tb]
\setlength{\tabcolsep}{4pt}
\centering
\small
\begin{tabular}{ l r r r r r r r r r r  }
\toprule
 Adaptation data & Fine-tuning & Dropout & Warmup & Stable & Learning & Weight & Batch & Epochs & Tuner & Avg \\
 & steps & & steps & & rate & decay & size &  & trials & (ppl) \\
\midrule
ByGPT5 base & --- & --- & --- & --- & --- & --- & --- & ---  & --- & 2.85  \\
\midrule
~~$\vsa$~C4 0.90           & 1 &  .458 & 760  & .096 & 2.77e-04 & .382 & 592  & 1 & 88  & 2.21 \\
~~~~~$\vsa$~Subtitle 0.75  & 2 &  .433 & 55   & .948 & 6.10e-08 & .112 & 2368 & 1 & 156 & 2.22 \\
~~~~~~~~$\vsa$~BOLT+Daily  & 3 & .567 & 142 & .109 & 6.07e-06 & .109 & 6720  &  2 &  105  &  2.14 \\    
\midrule
~~$\vsa$~BOLT+Daily        & 1 & .701 & 423 & .421 & 4.83e-05 & .514 &  5600 & 10  & 77 &  2.27 \\
\bottomrule
\end{tabular}
\caption{Hyperparameters after the specified number of tuning trials during our adaptation process of the ByGPT5 LLM. Also shown is the average perplexity on the dev sets. $\vsa$ denotes separate fine-tuning steps. + denotes datasets merged in a single fine-tuning step.\label{table_hyper_byte}}
\end{table*}


\begin{table*}[tb]
\setlength{\tabcolsep}{4pt}
\centering
\small
\begin{tabular}{ l r r r r r r r r r r  }
\toprule
 Training data           & Training & Dropout & Warmup & Stable & Learning & Weight & Batch & Examples per & Tuner  & Avg   \\
                           & steps        &         & steps  &        & rate     & decay  & size  & sentence     & trials & (ppl) \\
\midrule
~~$\vsa$~C4 0.90           & 1           & .583    & 238    & .453   & 1.08e-04 & .642   & 1120  & 1            & 115    & 2.64  \\
~~~~~$\vsa$~Subtitle 0.75  & 2           & .361    & 541    & .009   & 1.01e-06 & .438   & 6720  & 1            & 142    & 2.59  \\
~~~~~~~~$\vsa$~BOLT        & 3           & .322    & 955    & .807   & 2.15e-05 & .449   & 5600  & 5            & 164    & 2.55  \\    
~~~~~~~~~~~$\vsa$~Daily    & 4           & .725    & 340    & .569   & 3.42e-05 & .185   & 8960  & 4            & 1881   & 2.54  \\    
~~~~~~~~$\vsa$~BOLT+Daily  & 3           & .201    & 13     & .535   & 2.80e-05 & .379   & 3360  & 20           & 76     & 2.47  \\    
\midrule
~~$\vsa$~BOLT+Daily        & 1           & .011    & 431    & .492   & 1.19e-04 & .687   & 6720  & 47           & 50     & 2.60  \\
\bottomrule
\end{tabular}
\caption{Hyperparameters after the specified number of tuning trials during our adaptation process of the opt-350m model with a classification layer. Also shown is the average perplexity on the dev sets. $\vsa$ denotes separate training steps. + denotes datasets merged in a single training step.\label{table_hyper_classifier}}
\end{table*}

As an alternative to a subword LLM and our search algorithm, we tested using the ByGPT5 LLM \cite{belouadi_bygpt5}. ByGPT5 uses byte instead of subword tokenization. We fine-tuned ByGPT5 similarly to the subword opt-350m model (see \cref{appendix_subword_hyper}). Since fine-tuning took around four times as long compared to opt-350m, we used only one million sentences when tuning on C4 and Subtitle. This allowed the tuner to explore a similar number of parameter settings as opt-350m. For the byte model, the batch size was chosen from 592, 1184, 1776, 2368, 2960, 3552, 4144, or 4736 to maximize memory use during model fine-tuning. 

\cref{table_hyper_byte} shows the tuned hyperparameters and the resulting dev set perplexity at each step of the adaptation curriculum. Predictions improved markedly compared to the base model, likely due to the multilingual nature of the original model. We found that unlike the subword opt-350m model, the byte model benefited from first fine-tuning on C4 and then subsequently fine-tuning on Subtitle.

\subsection{Classifier model adaptation\label{appendix_class_results}}


We experimented with adding a classification layer to our domain-adapted version of opt-350m~\cite{zhang_opt}. \cref{table_hyper_classifier} (rightmost column) shows the perplexity on the dev sets at each step in the curriculum. Because the base classification model had already received the domain adaptation, the primary aim of this curriculum was to teach the model the classification task. We did not observe large perplexity changes between stages as with the byte model. However, perplexity did decrease slightly from step-to-step. Similar to during domain adaptation of opt-350m, we found combining BOLT and Daily in the final step improved performance. We found it helped to first train the classification model on C4 and then Subtitle compared to training on just a combination of BOLT and Daily.


We fine-tuned the opt-350m~\cite{zhang_opt} classifier model in the same way as the subword model (\cref{appendix_subword_hyper}). Since the training examples chose a location within each sentence (\cref{classifier_section}), we tuned the number of examples created from each sentence in place of epochs. For the C4 and Subtitle sets, this was fixed at 1. For BOLT, Daily, and BOLT+Daily, this was chosen from between 1 and 50, inclusive. \cref{table_hyper_classifier} shows the selected hyperparameters for each step in the curriculum.

\section{Search algorithm pseudocode}\label{appendix_pseudocode}
This work relies heavily on our search algorithm (\cref{sec_search_alg}). To convey the algorithm's structure more clearly, we produced pseudocode in addition to the Python code in our supplementary materials. Algorithm \ref{alg_vocab_hash} describes the vocab hash built at startup to make the search efficient. The hash maps each possible text sequence to valid matching tokens. Algorithm \ref{alg_search} describes the search algorithm used for each inference.

\begin{algorithm}
\small
    \DontPrintSemicolon
    \SetAlgoLined
    \KwData{model vocabulary, valid symbol set}
    \KwResult{initializes vocabulary hash}
    \BlankLine
    $valid\_vocab \leftarrow \emptyset$\;
    $vocab\_hash \leftarrow \emptyset$\;

    \ForEach{token, index $\in$ model\_vocab}{
        \If{each character in $token \in symbol\_set$}{
            \tcc{This is a valid token given our symbol set.}
            \emph{append $index$ to $valid\_vocab$}\;
            \tcc{Add this token index to the hash for every key that is a prefix of the token text.}
            \For{$i \leftarrow 0$ \KwTo len($token$)}{
                $key \leftarrow word[0$ \KwTo $i+1]$\;
                \emph{append $index$ to $vocab\_hash[key]$}\;
            }
        }
    }
    
    \caption{Initializes a hash of model tokens that begin with a given prefix.}
    \label{alg_vocab_hash}
\end{algorithm}

\begin{algorithm*}
\small
    \DontPrintSemicolon
    \SetAlgoLined
    \KwData{context, valid symbol set, beam width, max completed}
    \KwResult{probability distribution over symbol set}
    \BlankLine

    $target\_index \leftarrow$ len($context$)\;
    $last\_space \leftarrow$ last index of space in $context$\;
    $tokens \leftarrow $ tokenize($context[0$ \KwTo $last\_space]$)\;
    $current\_hypos, next\_hypos \leftarrow$ empty heap\;
    $char\_dict \leftarrow$ empty dictionary\;
    $completed \leftarrow 0$\;
    
    \tcc{Store hypotheses as triples of log prob, token sequence, and character length.}
    $initial\_hypo \leftarrow$ 0.0, $tokens$, $last\_space$\;
    \emph{push $initial\_hypo$ to $current\_hypos$}\;

    \While{$current\_hypos$ not empty}{
        \emph{sort $current\_hypos$ by descending log prob}\;
        $model\_input \leftarrow $ all token seq from $current\_hypos$\;
        $log\_probs \leftarrow$ log softmax($model\_output$)\;

        \ForEach{hypo, hypo\_index $\in$ current\_hypos}{
            \tcc{Determine what context needs to be matched based on hypothesis length.}
            $remaining \leftarrow context[hypo[length]$) \KwTo $end]$\;
            \tcc{Narrow the search to only tokens that match the remaining context.}
            \eIf{len($remaining$) $= 0$}{
                $search\_vocab \leftarrow valid\_vocab$\;
            }{
                $search\_vocab  \leftarrow vocab\_hash[remaining]$\;
                \For{$i \leftarrow 1$ \KwTo len($remaining$)}{
                    $prefix \leftarrow remaining[0$ \KwTo $i]$\;
                    \If{$prefix$ is single token}{
                        \emph{append tokenize($prefix$) to $search\_vocab$}\;
                    }
                }
            }

            \ForEach{$token \in search\_vocab$}{
                \eIf{$hypo[length]$) $+$ len($token$) $> target\_index$ }{
                    \tcc{This hypo has surpassed the context. Add its log prob to the list of log probs for the first character following the context.}
                    $char \leftarrow token[target\_index - hypo[length]$)$]$\;
                    $log\_prob \leftarrow hypo[log\_prob] + log\_probs[hypo\_index][token]$\;
                    \emph{append $log\_prob$ to $char\_dict[char]$}\;
                    \emph{increment $completed$}\;
                }{
                    \tcc{Add the extended hypothesis to the heap for the next round.}
                    $new\_hypo \leftarrow log\_prob, (hypo[tokens] + token), (hypo[length] + len(token))$\;

                    \tcc{Prune the list of hypotheses based on the beam width.}
                    \emph{push $next\_hypo$ to $next\_hypos$}\;
                    \If{count($next\_hypos$) $>$ $beam\_width$}{
                        \emph{pop hypo with lowest log prob from $next\_hypos$}\;
                    }
                }
            }

            \If{$completed >= max\_completed$}{
                \tcc{Completed maximum number of hypotheses.}
                \emph{break from outer while loop}\;
            }
        }

        $current\_hypos \leftarrow next\_hypos$\;
        $next\_hypos \leftarrow$ empty heap\;
        
    }
    $log\_probs \leftarrow$ empty dictionary\;
    \ForEach{$char \in symbol\_set$}{
        \eIf{$char \in char\_dict$}{
            $log\_probs[char] \leftarrow$ logsumexp($char\_dict[char]$)\;
        }{
            $log\_probs[char] \leftarrow -inf$\;
        }
    }

    \Return{softmax($log\_probs$)}\;
    
    \caption{Conducts a search using a subword model to produce a probability distribution for the next character.}
    \label{alg_search}
\end{algorithm*}

\end{document}